# Learning Probabilistic Relational Dynamics for Multiple Tasks

**Ashwin Deshpande**
MIT CSAIL
Cambridge, MA 02139
ashwind@mit.edu

**Brian Milch**
MIT CSAIL
Cambridge, MA 02139
milch@csail.mit.edu

**Luke S. Zettlemoyer**
MIT CSAIL
Cambridge, MA 02139
lsz@csail.mit.edu

**Leslie Pack Kaelbling**
MIT CSAIL
Cambridge, MA 02139
lpk@csail.mit.edu

## Abstract

The ways in which an agent's actions affect the world can often be modeled compactly using a set of relational probabilistic planning rules. This paper addresses the problem of learning such rule sets for multiple related tasks. We take a hierarchical Bayesian approach, in which the system learns a prior distribution over rule sets. We present a class of prior distributions parameterized by a *rule set prototype* that is stochastically modified to produce a task-specific rule set. We also describe a coordinate ascent algorithm that iteratively optimizes the task-specific rule sets and the prior distribution. Experiments using this algorithm show that transferring information from related tasks significantly reduces the amount of training data required to predict action effects in blocks-world domains.

## 1 Introduction

One of the most important types of knowledge for an intelligent agent is that which allows it to predict the effects of its actions. For instance, imagine a robot that performs the familiar task of retrieving items from cabinets in a kitchen. This robot needs to know that if it grips the knob on a cabinet door and pulls, the door will swing open; if it releases its grip when the cabinet is only slightly open, the door will probably swing shut; and if it releases its grip when the cabinet is open nearly 90 degrees, the door will probably stay open. Such knowledge can be encoded compactly as a set of *probabilistic planning rules* [Kushmerick *et al.*, 1995; Blum and Langford, 1999]. Each rule specifies a probability distribution over sets of changes that may occur in the world when an action is executed and certain preconditions hold. To represent domains concisely, the rules must be relational rather than propositional: for example, they must make statements about cabinets in general rather than individual cabinets.

Algorithms have been developed for learning relational probabilistic planning rules by observing the effects of actions [Pasula *et al.*, 2004; Zettlemoyer *et al.*, 2005]. But with current algorithms, if a robot learns planning rules for one kitchen and then moves to a new kitchen where its actions have slightly different effects (because, say, the cabinets are built differently), it must learn a new rule set from scratch. Current rule learning algorithms fail to capture an important aspect of human learning: the ability to transfer knowledge from one task to another. We address this transfer learning problem in this paper.

In statistics, the problem of transferring predictions across related data sets has been addressed with *hierarchical Bayesian models* [Lindley, 1971]. The first use of such models for the multi-task learning problem appears to be due to Baxter [1997]; the approach has recently become quite popular [Yu *et al.*, 2005; Marx *et al.*, 2005; Zhang *et al.*, 2006]. The basic idea of hierarchical Bayesian learning is to regard the task-specific models $R_1, \ldots, R_K$ as samples from a global prior distribution $G$. This prior distribution over models is not fixed in advance, but is learned by the system; thus, the system discovers what the task-specific models have in common.

However, applying the hierarchical Bayesian approach to sets of first-order probabilistic planning rules poses both conceptual and computational challenges. In most existing applications, the models $R_k$ are represented as real-valued parameter vectors, and the hypothesis space for $G$ is a class of priors over real vectors. But a rule set is a discrete structure that may contain any number of rules, and each rule includes a precondition and a set of outcomes that are represented as arbitrary-length conjunctions of first-order literals. How can we define a class of prior distributions over such rule sets? Our proposal is to let $G$ be defined by a *rule set prototype* that is modified stochastically to create the task-specific rule sets.

Our goal is to take data from $K$ *source tasks*, plus a limited set of examples from a *target task* $K+1$, and find the rule set $R^*_{K+1}$ for the target task with the greatest posterior probability. In principle, this involves integrating out the

$$\textit{pickup}(X,Y): \quad \textit{on}(X,Y), \textit{clear}(X), \textit{inhand-nil}, \textit{block}(Y), \neg\textit{wet}$$
$$\rightarrow \begin{cases} .7: & \textit{inhand}(X), \neg\textit{clear}(X), \neg\textit{inhand-nil}, \neg\textit{on}(X,Y), \textit{clear}(Y) \\ .2: & \textit{on}(X, \text{TABLE}), \neg\textit{on}(X,Y) \\ .05: & \text{no change} \\ .05: & \text{noise} \end{cases}$$

$$\textit{pickup}(X,Y): \quad \textit{on}(X,Y), \textit{clear}(X), \textit{inhand-nil}, \textit{block}(Y), \textit{wet}$$
$$\rightarrow \begin{cases} .2: & \textit{inhand}(X), \neg\textit{clear}(X), \neg\textit{inhand-nil}, \neg\textit{on}(X,Y), \textit{clear}(Y) \\ .2: & \textit{on}(X, \text{TABLE}), \neg\textit{on}(X,Y) \\ .3: & \text{no change} \\ .3: & \text{noise} \end{cases}$$

Figure 1: Two rules for the *pickup* action in the "slippery gripper" blocks world domain.

other rule sets $R_1, \ldots, R_K$ and the rule set prototype $G$. As an approximation, however, we use estimates of $G^*$ and $R_1^*, \ldots, R_K^*$ found by a greedy local search algorithm. We present experiments with this algorithm on blocks world tasks, showing that transferring data from related tasks significantly reduces the number of training examples required to achieve high accuracy on a new task.

## 2 Probabilistic Planning Rules

Probabilistic planning rule sets define a state transition distribution $p(s_t|s_{t-1}, a_t)$. In this section, we present a simplified version of the representation developed by [Zettlemoyer *et al.*, 2005]. A state $s_t$ is represented by a conjunctive formula with constants denoting objects in the world and proposition and function symbols representing the objects' properties and relations. The sentence

$$\textit{inhand-nil} \wedge \textit{on}(\text{B-A}, \text{B-B}) \wedge \textit{on}(\text{B-B}, \text{TABLE}) \wedge \textit{clear}(\text{B-A}) \wedge \textit{block}(\text{B-A}) \wedge \textit{block}(\text{B-B}) \wedge \textit{table}(\text{TABLE}) \quad (1)$$

represents a blocks world where the gripper holds nothing and the two blocks are in a single stack on the table. This is a full description of the world; all of the false literals are omitted for compactness. Block B-A is on top of the stack, while B-B is below B-A and on the table TABLE. Actions $a_t$ are ground literals where the predicate names the action to be performed and the arguments are constant terms that correspond to the objects which will be manipulated. For example, $a_t = \textit{pickup}(\text{B-A}, \text{B-B})$ would represent an attempt to pick block B-A up off of block B-B.

Each rule $r$ has two parts that determine when it is applicable: an action $z$ and a context $\Psi$ that encodes a set of preconditions. Both of the rules in Fig. 1 model the *pickup*$(X, Y)$ action. Given a particular state $s_{t-1}$ and action $a$, we can determine whether a rule *applies* by computing a binding $\theta$ that finds objects for all the variables, by matching against $a$, and then testing whether the preconditions hold for this binding. For example, for the state $s$ in sentence 1 and $a = \textit{pickup}(\text{B-A}, \text{B-B})$, both of the rules in Fig. 1 would have the binding $\theta = \{X/\text{B-A}, Y/\text{B-B}\}$. The first rule would apply, since its preconditions are all satisfied, while the second one would not because *wet* is not true in $s$. We disallow rule sets in which two or more rules apply to the same $(s, a)$ pair (these are called *overlapping rules*). In cases where no rules apply, a default rule is used that has an empty context and two outcomes: no change and noise, which will be described shortly.

Given the applicable rule $r$, the discrete distribution $\mathbf{p}$ over outcomes $O$, described on the right of the $\rightarrow$, defines what changes may happen from $s_{t-1}$ to $s_t$. Each non-noise outcome $o \in O$ implicitly defines a *successor state function* $f_o$ with associated probability $p_o$, an entry in $\mathbf{p}$. The function $f_o$ builds $s_t$ from $s_{t-1}$ by copying $s_{t-1}$ and then changing the values of the relevant literals in $s_t$ to match the corresponding values in $\theta(o)$. In our running example of executing *pickup*(B-A, B-B) in sentence 1, for the first outcome of the first rule, where the picking up succeeds, $f_o$ would set five truth values, including setting *on*(B-A, B-B) to be false. In the third outcome, which indicates no change, $f_o$ is the identity function. In this paper, we will enforce the restriction that outcomes do not overlap: for each pair of outcomes $o_1$ and $o_2$ in a rule $r$, there cannot exist a state–action pair $(s, a)$ such that $r$ is applicable and $f_{o_1}(s) = f_{o_2}(s)$. In other words, if we observe the state that results from applying a rule, then there is no ambiguity about which outcome occurred.[1] Finally, the *noise outcome* is treated as a special case. There is no associated successor function, which allows the rule to define a type of partial model where $r$ does not describe how to construct the next state with probability $p_{noise}$. Noise outcomes allow rule learners to ignore overly complex, rare action effects and have been shown to improve learning in noisy domains [Zettlemoyer *et al.*, 2005]. Since rules with noise outcomes are partial models, the distribution $p(s_t|s_{t-1}, a_t)$ is replaced with an approximation:

$$\hat{p}(s_t|s_{t-1}, a_t) = \begin{cases} p_o & \text{if } f_o(s_{t-1}) = s_t \\ p_{noise}p_{min} & \text{otherwise} \end{cases} \quad (2)$$

where the set of possible outcomes $o \in O$ is determined by the applicable rule. The probabilities $p_o$ and $p_{noise}$ make up the parameter vector $\mathbf{p}$. The constant $p_{min}$ can be viewed as an approximation to a distribution $p(s_t|s_{t-1}, a_t, o_{noise})$ that would provide a complete model.

## 3 Hierarchical Bayesian Model

In a hierarchical Bayesian model, as illustrated in Fig. 2, the data points $x_{kn}$ in task $k$ come from a task-specific dis-tribution $p(x_{kn}|R_k)$, and the task-specific parameters $R_k$ are in turn modeled by a prior distribution $p(R_k|G)$. The hyperparameter $G$ has its own prior distribution $p(G)$. By observing data from the first $K$ tasks, the learner gets information about $R_1, \ldots, R_K$ and hence about $G$. For instance, the learner can compute (perhaps approximately) the values $(R_1^*, \ldots, R_K^*, G^*)$ that have maximum *a posteriori* (MAP) probability given the data on the first $K$ tasks. Then when it encounters task $K+1$, the learner's estimates of the task-specific model $R_{K+1}$ are influenced by both the data observed for task $K+1$ and the prior $p(R_{K+1}|G^*)$, which captures its expectations about the model based on the preceding tasks.

[1] This restriction simplifies parameter estimation (as we will see in Sec. 4) without limiting the class of transition distributions that can be defined. Any rule with overlapping outcomes can be replaced by an equivalent set of rules applying to more specific contexts, with non-overlapping outcomes.

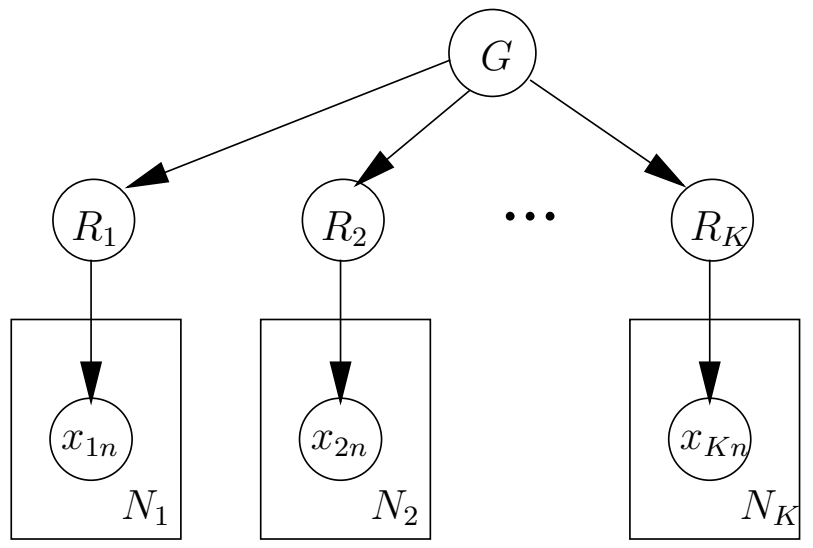

Figure 2: A hierarchical Bayesian model with $K$ tasks, where the number of examples for task $k$ is $N_k$.

### 3.1 Rule Set Prototypes

In the context of learning planning rules, the task-specific models $R_k$ are rule sets. Our intuition is that if the tasks are related, then these rule sets have some things in common. Certain rules may appear in the rule sets for many tasks, perhaps with some modifications to their contexts, outcomes, and outcome probabilities. To capture these commonalities, we assume that the rule sets are all generated from an underlying *rule set prototype* $G$.

A rule set prototype consists of a set of *rule prototypes*. A rule prototype is like an ordinary rule, except that rather than specifying a probability distribution over its outcomes, it specifies a vector of Dirichlet parameters that define a prior over outcome distributions. For a rule prototype with $n$ explicit outcomes, this is a vector $\Phi$ of $n+2$ non-negative real numbers: $\Phi_{n+1}$ corresponds to a special *seed outcome* $o_{n+1}^*$ that generates new outcomes in local rules, and $\Phi_{n+2}$ accounts for the noise outcome. Unlike in local rule sets, we allow overlapping rules and outcomes in rule set prototype to allow for better generalization.

### 3.2 Overview of Model

Our hierarchical model defines a joint probability distribution $p(G, R_1, \ldots, R_K, x_1, \ldots, x_K)$. In our setting, each example $x_{kn}$ is a state $s_t$ obtained by performing a known action $a_t$ in a known initial state $s_{t-1}$. Thus, $p(x_{kn}|R_k)$ can be found by identifying the single rule in $R_k$ that applies to $(s_{t-1}, a_t)$ (or the default rule, if no explicit rule applies) and using Eq. 2. Then the probability of the entire data set for task $k$ is $p(x_k|R_k) = \prod_{n=1}^{N_k} p(x_{kn}|R_k)$.

The distribution for $G$ and $R_1, \ldots, R_k$ is defined by a generative process that first creates $G$, and then creates $R_1, \ldots, R_k$ by modifying $G$. Note that this generative process is purely a conceptual device for defining our probability model: we never actually draw samples from it. As we will see in Sec. 4, our learning algorithm uses the generative model solely to define a scoring function for evaluating rule sets and prototypes.

Two difficulties arise in using our generative process to define a joint distribution. One is that the process can yield rule sets $R_i$ that are invalid, in the sense of containing overlapping rules or outcomes. It is difficult to design a generative process that avoids creating invalid rule sets, but still allows the probability of a rule set to be computed efficiently. Intuitively, we want to discard runs of the generative process that yield invalid rule sets. The other difficulty is that there may be many possible runs of a generative process that yield the same rule set. For instance, as we will see, a rule set prototype is generated by choosing a number $m$, generating a sequence of $m$ rule prototypes independently, and then returning the set of distinct rule prototypes that were generated. In principle, a set of $m^*$ distinct rules could be created by generating a list of any length $m \geq m^*$ (with duplicates); we do not want to force ourselves to sum over all these possibilities to compute the probability of a given rule set prototype. Again, it is convenient to discard certain non-canonical runs of the generative process: in this case, runs where the same rule prototype is generated twice.

Thus, we will define measures $P_G(G)$ and $P_{\text{mod}}(R_k|G)$ that give the probability of generating a rule set prototype $G$, or a rule set $R_k$, through a "valid" sampling run. Because some runs are considered invalid, these measures do not sum to one. The resulting joint distribution is:

$$p(G, R_1, \ldots, R_K, x_1, \ldots, x_K) = \frac{1}{Z} P_G(G) \prod_{k=1}^{K} P_{\text{mod}}(R_k|G) p(x_k|R_k) \quad (3)$$

The normalization constant $Z$ is the total probability of valid runs of our generative process. Since we are just interested in the relative probabilities of hypotheses, we never need to compute this normalization constant.[2]

[2]One might be tempted to define a model where the normalization is more local: for instance, to replace the factor $P_{\text{mod}}(R_k|G)$ in Eq. 3 with a normalized distribution $P_{\text{mod}}(R_k|G)/Z(G)$. However, the normalization factor $Z(G)$ is not constant, so it would have to be computed to compare alternative values of $G$.

### 3.3 Modifying the Rule Set Prototype

We begin the discussion of our generative process by describing how a rule set prototype $G$ is modified to create a rule set $R$ (the process that generates $G$ will be a simplified version of this process). The first step is to choose the rule set size $m$ from a distribution $P_{\text{num}}(m|m^*)$, where $m^*$ is the number of rule prototypes in $G$. We define $P_{\text{num}}(m|m^*)$ so that all natural numbers have non-zero probability, but $m$ is likely to be close to $m^*$, and the probability drops off geometrically for greater values of $m$.

$$P_{\text{num}}(m|m^*) = \begin{cases} \text{Geom}[\alpha](m-m^*) & \text{if } m > m^* \\ (1-\alpha)\text{Binom}[m^*,\beta](m) & \text{otherwise} \end{cases} \quad (4)$$

Here $\text{Geom}[\alpha]$ is a geometric distribution with success probability $\alpha$. Thus, if $m > m^*$, then $P_{\text{num}}(m|m^*) = (1-\alpha)\alpha^{(m-m^*)}$. We set $\alpha$ to a small value to discourage the rule set $R$ from being much larger than $G$. The sum of the $\text{Geom}[\alpha]$ distribution over all values greater than zero is $\alpha$, leaving a probability mass of $1-\alpha$ to be apportioned over rule set sizes from 0 through $m^*$. The binomial distribution $\text{Binom}[m^*,\beta]$ — which yields the probability of getting exactly $m$ heads when flipping $m^*$ coins with heads probability $\beta$ — is a convenient distribution over this range of integers. We set $\beta$ to a value close to 1 to express a preference for local rule sets that are not much smaller than the prototype set.

Next, for $i = 1$ to $m$, we generate a local rule $r_i$. The first step in generating $r_i$ is to choose which rule prototype in $G$ it will be derived from. This choice is represented by an assignment variable $A_i$, whose value is either a rule prototype in $G$, or a special value NIL indicating that this rule is generated from scratch with no prototype. The distribution $P_A(a_i|G)$ assigns the probability $\gamma_{\text{rule}}$ to NIL and spreads the remaining mass uniformly over the rule prototypes. Since the $A_i$ are chosen independently, a single rule in $G$ may serve as the prototype for several rules in $R$, or for none. Next, given the rule prototype (or null value) $a_i$, the local rule $r_i$ is generated according to a distribution $P_{\text{rule}}(r_i|a_i)$. We discuss this distribution in Section 3.4.

The rule set generated by this process is the set of distinct rules in the list $r_1, \ldots, r_m$. We consider a run of the generative process to be invalid if any of these rules have overlapping contexts; in particular, this constraint rules out cases where the same rule occurs twice. So the probability of generating a set $\{r_1, \ldots, r_m\}$ on a valid run is the sum of the probabilities of all permutations of this set. This is $m!$ times the probability of generating the rules in any particular order. Thus, the probability of getting a valid local rule set $R$ of size $m$ from a prototype $G$ of size $m^*$ is:

$$P_{\text{mod}}(R|G) = P_{\text{num}}(m|m^*) \cdot m! \cdot \prod_{i=1}^{m} \sum_{\substack{a_i \in \\ G \cup \{\text{NIL}\}}} P_A(a_i|G) P_{\text{rule}}(r_i|a_i) \quad (5)$$

### 3.4 Modifying and Creating Rules

We will now define the distribution $P_{\text{rule}}(r|r^*)$, where $r^*$ may be either a rule prototype, or the value NIL, indicating that $r$ is generated from scratch. Suppose $r$ consists of a context formula $\Psi$, an action term $z$, a set of non-noise outcomes $O$, and a probability vector $\mathbf{p}$. The corresponding parts of $r^*$ will be referred to as $\Psi^*$, $z^*$, $O^*$, and $\Phi$ (recall that this last component is a vector of Dirichlet parameters). If $r^* = \text{NIL}$, then $\Psi^*$ is an empty formula, $z^*$ is NIL, $O^*$ consists of just the seed outcome, and $\Phi$ is a two-element vector consisting of a 1 for the seed outcome and a 1 for the noise outcome.

For rules derived from a rule prototype, we assume the action term is unchanged. So if $z^*$ is not NIL, we use the distribution $P_{\text{act}}(z|z^*)$ that assigns probability one to $z^*$. If a rule is generated from scratch, we need to generate its action term. For simplicity, we assume that each action term consists of an action symbol and a distinct logical variable for each argument; we do not allow repeated variables or more complex terms in the argument list. The distribution $P_{\text{act}}(z|z^*)$ chooses the action term uniformly from the set of such terms when $z^* = \text{NIL}$.

The next step in generating $r$ is to choose its context $\Psi$. We define the distribution for $\Psi$ by means of a general formula-modification distribution $P_{\text{for}}(\Psi|\Psi^*, \bar{v})$, where $\bar{v}$ is the set of logical variables that occur in $z$ and thus are eligible to be included in $\Psi$. This distribution is explained in Sec. 3.5.

To generate the outcome set $O$ from $O^*$, we use essentially the same method we used to generate the rule set $R$ from $G$. We begin by choosing the size $n$ of the outcome set from the distribution $P_{\text{num}}(n|n^*)$, where $n^* = |O^*|$. The distribution $P_{\text{num}}$ here is the same one used in Sec. 3.3 (one could use different $\alpha$ and $\beta$ parameters here). Then, for $i = 1$ to $n$, we choose which prototype outcome serves as the source for the $i$th local outcome. This choice is represented by an assignment variable $B_i$. As in the case of rules, we allow some local outcomes to be generated from scratch rather than from a prototype; this choice is represented by the seed outcome. The value of $B_i$ is chosen from $P_B(b_i|O^*)$, which assigns probability $\gamma_{\text{out}}$ to the seed outcome and is uniform over the rest of the outcomes.

Once the source for each local outcome has been chosen, the next step is to generate the outcomes themselves. Recall that an outcome is just a formula. Thus, we define the outcome modification distribution using the general formula-modification process $P_{\text{for}}(o_i|b_i, \bar{v})$ that we will discuss in Sec. 3.5 (again, $\bar{v}$ is the set of logical variables in $z$). If $b_i$ is the seed outcome, then $P_{\text{for}}$ treats it as an empty formula. A list of outcomes is considered valid if it contains no repeats and no overlapping outcomes. Since repeats are excluded, the probability of a set of $n$ outcomes is $n!$ times the probability of any corresponding list. Thus, we get the following probability of generating a valid outcome set $O$ and an assignment vector $\mathbf{b}$, given that the prototype outcome set is $O^*$ and the number of local outcomes is $n$:

$$P_{\text{out}}(O, \mathbf{b}|O^*, n) = n! \prod_{i=1}^{n} P_B(b_i|O^*) P_{\text{for}}(o_i|b_i, \bar{v}) \tag{6}$$

The last step is to generate the outcome probabilities $\mathbf{p}$. These probabilities are sampled from a Dirichlet distribution whose parameter vector depends on the prototype parameters $\Phi$ and the assignment vector $\mathbf{b} \equiv (b_1, \ldots, b_n)$. Specifically, define the function $f(\Phi, \mathbf{b})$ to yield a parameter vector $(\Phi'_1, \ldots, \Phi'_{n+1})$ such that:

$$\Phi'_i = \begin{cases} \frac{\Phi_{b_i}}{C(\mathbf{b}, b_i)} & \text{if } i \leq n \\ \Phi_{n+2} & \text{if } i = n+1 \end{cases} \tag{7}$$

This definition says that if $o_i$ is generated from prototype outcome $b_i$ (including the seed outcome), then $\Phi'_i$ is obtained by dividing up $\Phi_{b_i}$ over all the local outcomes derived from $b_i$. The number of such outcomes is computed by the function $C(\mathbf{b}, b_i)$, which returns the number of indices $j \in \{1, \ldots, n\}$ such that $b_j = b_i$. Finally, for the noise outcome, we have $\Phi'_{n+1} = \Phi_{n+2}$.

To define the overall distribution for a local rule $r$ given a rule prototype $r^*$, we sum out the assignment variables $B_i$. For valid rules $r$, we get:

$$P_{\text{rule}}(r|r^*) = P_{\text{act}}(z|z^*)\, P_{\text{for}}(\Psi|\Psi^*, \bar{v})\, P_{\text{num}}(n|n^*) \cdot \sum_{\substack{\mathbf{b} \in \\ (O^* \cup \{\text{NIL}\})^n}} P_{\text{out}}(O, \mathbf{b}|O^*, n)\, \text{Dir}[f(\Phi, \mathbf{b})](\mathbf{p}) \tag{8}$$

Here $\text{Dir}[f(\Phi, \mathbf{b})]$ is the Dirichlet distribution with parameter vector $f(\Phi, \mathbf{b})$.

### 3.5 Modifying Formulas

The formulas that serve as contexts and outcomes are very simple: they are just conjunctions of literals, where a literal has the form $t = x$ for some term $t$ and value $x$. The term must be *simple* in the sense that each of its arguments is either a constant symbol or a logical variable; similarly, $x$ must be a constant symbol or a logical variable.[3] We do not care about the order of literals in a formula, and we would also like to rule out self-contradictory formulas in which multiple values are assigned to the same term. It is convenient to think of a formula $\varphi$ as a pair $(T, I)$, where $T$ is a set of simple terms and $I$ is a function from elements of $T$ to values. This representation guarantees that the elements of $T$ are unordered, and each element is mapped to only one value.

[3] We are treating *true* and *false* as constant symbols, so a literal such as $\neg on(X, Y)$ is represented as $on(X, Y) = \textit{false}$.

So to define our formula-modification distribution $P_{\text{for}}(\varphi|\varphi^*, \bar{v})$, we will suppose $\varphi = (T, I)$ and $\varphi^* = (T^*, I^*)$. Recall that $\bar{v}$ is the set of logical variables that may be used in $\varphi$ and $\varphi^*$. To generate $\varphi$, we first choose a set $T_{\text{keep}} \subseteq T^*$, where each term in $T^*$ is included in $T_{\text{keep}}$ independently with probability $\beta_{\text{term}}$. The terms in $T_{\text{keep}}$ will be included in $T$. Next, we generate a set $T_{\text{new}}$ of new terms to include in $T$. The size of $T_{\text{new}}$, denoted $k_{\text{new}}$, is chosen from a geometric distribution with parameter $\alpha_{\text{term}}$. Then, for $i = 1$ to $k_{\text{new}}$, we generate a term $t_i$ according to a distribution $P_{\text{term}}(t_i|\bar{v})$. This distribution chooses a predicate or function symbol $f$ uniformly at random, and then chooses each argument of $f$ uniformly from the set of constant symbols plus $\bar{v}$. We consider a run invalid if any element of $T_{\text{new}}$ is in $T^*$: this ensures that while computing the probability of a term set $T$ given a prototype term set $T^*$, we can recover $T_{\text{keep}}$ as $T \cap T^*$ and $T_{\text{new}}$ as $T \setminus T^*$.

Next, we choose the term-to-value function $I$. For a term $t \in T \cap T^*$, the value $I(t)$ is equal to $I^*(t)$ with probability $\rho$, and with probability $(1 - \rho)$ it is sampled according to a distribution $P_{\text{value}}(x|\bar{v})$. If $t \notin T^*$, then $I(t)$ is always sampled from $P_{\text{value}}(x|\bar{v})$. This distribution $P_{\text{value}}(x|\bar{v})$ is uniform over the constant symbols in the language, plus $\bar{v}$.

### 3.6 Generative Model for Rule Set Prototypes

The process that generates rule set prototypes $G$ is similar to the process that generates local rule sets from $G$, but all the rule prototypes are generated from scratch — there are no higher-level prototypes from which they could be derived. We assume that the number of rule prototypes in $G$ has a geometric distribution with parameter $\alpha_{\text{proto}}$. Thus the probability of a rule set prototype $G$ of size $m^*$ with rule prototypes $\{r^*_1, \ldots, r^*_{m^*}\}$ is:

$$P_G(G) = \text{Geom}[\alpha_{\text{proto}}](m^*) \cdot m^*! \cdot \prod_{i=1}^{m^*} P_{\text{proto}}(r^*_i) \tag{9}$$

We consider a generative run to be invalid if it generates the same rule prototype more than once, although we allow rule prototypes to have overlapping contexts.

The rule prototypes are generated independently from the distribution $P_{\text{proto}}(r^*)$. This is similar to the distribution for generating a local rule from scratch (as given by Eq. 8). The action term $z^*$ is chosen from the uniform distribution $P_{\text{act}}(z^*|\text{NIL})$; the context formula $\Psi^*$ is generated by running our formula modification process on the empty formula $\emptyset$ given the logical variables $\bar{v}$ from $z^*$; the number of outcomes $n^*$ has a geometric distribution; and each outcome $o^*$ in the outcome set $O^*$ is also generated from $P_{\text{for}}(o^*|\emptyset, \bar{v})$. The main difference from the case of local rules is that rather than generating an outcome probability vector $\mathbf{p}$, we generate a vector of Dirichlet weights $\Phi$, defining a prior over outcome distributions. We use a hyperprior $P_\Phi(\Phi|n^*)$ on $\Phi$ in which the sum of the Dirichlet weights has an exponential distribution. Thus, if $r^*$ consists of an action term $z^*$ containing logical variables $\bar{v}$, a context $\Psi^*$, and an outcome set $O^*$ of size $n^*$, then:

$$P_{\text{proto}}(r^*) = P_{\text{act}}(z^*|\text{NIL})\, P_{\text{for}}(\Psi^*|\emptyset, \bar{v}) \cdot \text{Geom}[\alpha](n^*) P_\Phi(\Phi|n^*) \prod_{o \in O^*} P_{\text{for}}(o|\emptyset, \bar{v})$$

# 4 Learning

In our problem formulation, we are given sets of examples $x_1, ..., x_K$ from $K$ source tasks, and a set of examples $(x_{K+1})$ from the target task . In principle, one could maximize the objective in Eq. 3 using the data from the source and target tasks simultaneously. However, if $K$ is fairly large, the data from task $K+1$ is unlikely to have a large effect on our beliefs about the rule set prototype $G$. Thus, we work in two stages. First, we find the best rule set prototype $G^*$ given the data for the $K$ source tasks. Then, holding $G^*$ fixed, we find the best rule set $R^*_{K+1}$ given $G^*$ and $x_{K+1}$. This approach has the benefit of allowing us to throw away our data from the source tasks, and just transfer the relatively small $G^*$.

Our goal in the first stage, then, is to find the prototype $G^*$ with the greatest posterior probability given $x_1, \ldots, x_K$. Doing this exactly would involve integrating out the source rule sets $R_1, \ldots, R_K$. It turns out that if we think of each rule set $R_k$ as consisting of a structure $R_k^{\text{S}}$ and parameters $R_k^{\text{P}}$ (namely the outcome probability vectors for all the rules), then we can integrate out $R_k^{\text{P}}$ efficiently. However, summing over all the discrete structures $R_k^{\text{S}}$ is difficult. Thus, we apply another MAP approximation, searching for the prototype $G$ and rule set structures $R_1^{\text{S}}, \ldots, R_K^{\text{S}}$ that together have maximal posterior probability. It is important that we integrate out the parameters $R_k^{\text{P}}$, because the posterior density for $R_k^{\text{P}}$ is defined over a union of spaces of different dimensions (corresponding to different numbers of rules and outcomes in $R_k$). The heights of density peaks in spaces of differing dimension are not necessarily comparable. So it would not be correct to use a MAP estimate of $R_k^{\text{P}}$ obtained by maximizing this density.

## 4.1 Scoring Function

In our search over $G$ and $R_1^{\text{S}}, \ldots, R_K^{\text{S}}$, our goal is to maximize the marginal probability obtained by integrating out the outcome probabilities:

$$P(G, R_1^{\text{S}}, \ldots, R_K^{\text{S}}) \propto P_G(G) \prod_{k=1}^{K} \int_{R_k^{\text{P}}} P_{\text{mod}}(R_k|G) P(x_k|R_k) \quad (10)$$

This equation trades off three factors: the complexity of the rule set prototype, represented by $P_G(G)$; the differences between the local rule sets and the prototype, $P_{\text{mod}}(R_k|G)$, and how well the local rule sets fit the data, $P(x_k|R_k)$.

Computing the value of Eq. 10 for a given choice of $G$ and $R_1, \ldots, R_K$ is expensive, because it involves summing over all possible mappings from local rules to global rules (the $a$ values in Eq. 5) and all mappings from local outcomes to prototype outcomes (the $b$ values in Eq. 8). Integrating out the outcome probabilities $\mathbf{p}$ in each rule is not a computational bottleneck: we can push the integral inside the sums over $a$ and $b$, and use a modified version of a standard estimation technique [Minka, 2003] for the Polya (or Dirichlet-multinomial) parameters.[4]

Rather than summing over all possible local-to-global correspondences for rules and outcomes, we approximate by using a single correspondence. Specifically, for each rule set $R_k \equiv \{r_1, \ldots, r_m\}$, we choose the rule correspondence vector $\hat{\mathbf{a}}$ that maximizes the probability of the local rule contexts $\Psi_i$ given the global rule contexts $\Psi_{(a_i)}$ (ignoring outcomes) $\hat{\mathbf{a}} = \text{argmax}_{\mathbf{a}} \prod_{i=1}^{m} P_A(a_i|G) P_{\text{for}}(\Psi_i|\Psi^*_{(a_i)}, \bar{v}_i)$. Since each factor contains only one assignment variable $a_i$, we can find the corresponding rule prototype for each local rule separately. Given the rule correspondence $\hat{\mathbf{a}}$, we next construct an outcome correspondence for each rule $r_i$. We use the outcome correspondence that maximizes the probability of the local outcomes $o_1, \ldots, o_n$ given the outcome set $O^*$ of the rule prototype $\hat{a}_i$ (ignoring the outcome probabilities) $\hat{\mathbf{b}} = \text{argmax}_{\mathbf{b}} \prod_{i=1}^{n} P_B(b_i|O^*) P_{\text{for}}(o_i|b_i, \bar{v})$. Again, the maximization decomposes into a separate maximization for each outcome. This greedy matching scheme can yield a poor result if a local rule $r_i$ has a context similar to a prototype rule, but very different outcomes. So as a final step, we compute the probability of each $r_i$ being generated from scratch, and set $\hat{a}_i$ to NIL if this is a better correspondence.

These approximations yield the following scoring function (an approximate version of Eq. 10), which we use to guide our search.

$$\mathit{Score}(G, R_1^{\text{S}}, \ldots, R_K^{\text{S}}) = P_G(G) \prod_{k=1}^{K} \int_{R_k^{\text{P}}} \widehat{P}_{\text{mod}}(R_k|G) P(x_k|R_k) \quad (11)$$

[4]We modify the standard technique to take into account our hyperprior $P_\Phi$. Also, we adjust for cases where some global outcomes are not included in a corresponding local rule. For a more detailed explanation, see the master's thesis by Deshpande [2007].

Here $\widehat{P}_{\mathrm{mod}}$ is a version of the measure $P_{\mathrm{mod}}$ from Eq. 5 in which we simply use $\hat{\mathbf{a}}$ rather than summing over $a_i$ values, and we replace $P_{\mathrm{rule}}$ with a modified version that uses $\hat{\mathbf{b}}$ rather than summing over $\mathbf{b}$ vectors.

### 4.2 Coordinate Ascent

We find a local maximum of Eq. 3 using a coordinate ascent algorithm. We alternate between maximizing over local rule set structures given an estimate of the rule set prototype $G$, and maximizing over the rule set prototype given estimates of the rule set structures $(R_1^{\mathrm{S}}, ..., R_K^{\mathrm{S}})$:

$$argmax_{R_1^{\mathrm{S}},\ldots,R_K^{\mathrm{S}}} \prod_{k=1}^{K} \int_{R_k^{\mathrm{P}}} P(x_k|R_k)P(R_k|G)$$

$$argmax_G \, P(G) \prod_{k=1}^{K} P(R_k^{\mathrm{S}}|G)$$

We begin with an empty rule set prototype, and use a greedy local search algorithm (described below) to optimize the local rule sets. Since $R_1, \ldots, R_K$ are conditionally independent given $G$, we can do this search for each task separately. When these searches stabilize — that is, no search operator improves the objective function — we run another greedy local search to optimize $G$. We repeat this alternation until no more changes occur.

### 4.3 Learning Local Rule Sets

During the coordinate ascent one task is to find the highest scoring local rule set $R_k^*$ given the rule set $G$. The search is closely related the rule set learning algorithm problem in Zettlemoyer *et al.* [2005]. There are three major differences: (1) $G$ provides a prior that did not exist before; (2) the outcomes $O$ for each rule are constrained to be non-overlapping; and (3) the rule parameters $\mathbf{p}$ are integrated out instead of being set to maximum likelihood estimates.

#### 4.3.1 Rule Set Search

In this section, we briefly outline a local rule learning algorithm that is a direct adaptation of the approach of Zettlemoyer *et al.* [2005] and highlight the places where the two algorithms differ. The search starts with a rule set that contains only the noisy default rule. At every step, we take the current rule set and apply a set of search operators to create new rule sets. Each of these new rule sets is scored, as described in section 4.1. The highest scoring set is selected and set as the new $R_k$, and the search continues until no new improvements are found.

The operators create new rule sets by directly manipulating the current set: either adding or removing some number of the existing rules. Whenever a new rule is created, the relevant operator constructs the rule's action and context and uses a subalgorithm to find the best set of outcomes. This outcome learning is done with a greedy search algorithm, as described in the next section. The following operators construct changes to the current rule set.

**Add/Remove Rule.** Two types of new rules can be added to the set. Rules can be created by an *ExplainExamples* procedure [Zettlemoyer *et al.*, 2005] which uses a heuristic search to find high quality potential rules in a data driven manner. In addition, rules can be created by copying the action and context of one of the prototypes in the global rule set. This provides a strong search bias towards rules that have been found to be useful for other tasks. New rule sets can also be created by removing one of the existing rules in the current set.

**Add/Remove Literal.** This operator selects a rule in the current rule set, and replaces it with a new rule that is the same except that one literal is added or removed from the context. All possible additions and deletions are proposed.

**Split on Literal.** This operator chooses an existing rule and a new term that does not occur in that rule's context. It removes the chosen rule and adds multiple new rules, one for each possible assignment of a value to the chosen term.

Any time a new rule is added to a rule set, there is a check to make sure that only one rule is applicable for each training example. Any preexisting rules with overlapping applicability are removed from the rule set.

#### 4.3.2 Outcome Search

Given a rule action $z$ and a context $\Psi$, the set of outcomes $O$ is learned with a greedy search that optimizes the score, computed as described in section 4.1. This algorithm is a modified version of a previous outcome search procedure [Pasula *et al.*, 2004], which has been changed to ensure that the outcomes do not overlap. Initially, $O$ contains only the noise outcome, which can never be removed. It each step, a set of search operators is applied to build new outcome sets, which are scored and the best one is selected. The search finishes when no improvements can be found. The operators include:

**Add/Remove Outcome.** This operator adds or removes an outcome from the set. Possible additions include any outcomes from the corresponding prototype rule or an outcome derived from concatenating the changes seen as a result of action effects in a training example (following [Pasula *et al.*, 2004]). Any existing outcome can be removed.

**Add/Remove Literal.** This operator appends or removes a literal from a specific outcome in the set. Any literal that is not present can be added and any currently present literal can be removed.

**Split on Literal.** This operator takes an existing outcome and replaces it with multiple new outcomes, each containing one of the possible value assignments for a new term.

**Merge Outcomes.** This operator creates a new outcome computing the union of an existing outcome and one that could be added by the add operator described above. The original outcome is removed from the set.

Two of the operators, add outcome and remove function, have the potential to create overlapping outcomes. To fix this condition, functions are greedily added to overlapping outcomes until no pair of outcomes overlap. This new outcome set is scored, and the search continues.

### 4.4 Learning the Rule Set Prototype

The second optimization involves finding the highest scoring rule set prototype $G$ given rule sets $(R_1^*, ..., R_K^*)$. Again, we adopt an approach based on greedy search through the space of possible rule sets. This search has exactly the same initialization and uses all of the same search operators as the local rule set search. There are three differences: (1) the *AddRule* operator tries to add rules that are present in the local rule sets, without directly referencing the training sets; (2) we relax the restriction that rules and outcomes can not overlap, simplifying some of the checking that the operators have to perform; and (3) we need to estimate the Dirichlet parameters for the outcomes for each new prototype rule considered by the structure search.

Estimating the Dirichlet parameters for the Polya distribution does not have a closed form solution, but gradient ascent techniques have been developed for the maximum likelihood solution [Minka, 2003]. To estimate the parameters for a rule prototype $r^*$, the required occurrence counts are computed for each prototype outcome and each local rule that corresponds to $r^*$ (under the correspondence $\hat{\mathbf{a}}$ described in Sec. 4.1). If a local rule contains several outcomes corresponding to the same prototype outcome (under $\hat{\mathbf{b}}$), their counts are merged.

## 5 Experiments

We evaluate our learning algorithm on synthetic data from four families of related tasks, all variants of the classic blocks world. We restrict ourselves to learning the effects of a single action, *pickup*$(X, Y)$. Adding more actions would not significantly change the problem: since the action is always observed, one can learn a rule set for multiple actions by learning a rule set for each action separately.

### 5.1 Methodology

Each run of our experiments consists of the following steps:

1. Generate $K$ "source task" rule sets from a prior distribution. This prior distribution is implemented by a special-purpose program for each family of tasks. This is slightly more realistic than generating the rule sets from a rule set prototype expressed in our modeling language.

2. For each source task, generate a set of $N_{\text{source}}$ state transitions to serve as a training set. In each state transition, the action is *pickup*$(\text{A}, \text{B})$ and the initial state is created by assigning random values to all functions on $\{\text{A}, \text{B}\}$.[5] Then the resulting state is sampled according to the task-specific rule set. Note that the state transitions are sampled independently of each other; they do not form a trajectory.

3. Run our full learning algorithm on the $K$ source-task training sets to find the best rule set prototype $G^*$.

4. Generate a "target task" rule set $R_{K+1}$ from the same distribution used in Step 1.

5. Generate a training set of $N_{\text{target}}$ state transitions as in Step 2, using $R_{K+1}$ as the rule set.

6. Learn a rule set $\widehat{R}_{K+1}$ for the target task using the algorithm from Sec. 4.3, with $G^*$ as the fixed rule set prototype.

7. Generate a test set of 1000 initial states using the same distribution as in Step 2. For each initial state $s$, compute the *variational distance* between the next-state distributions defined by the true rule set $R_{K+1}$ and the learned rule set $\widehat{R}_{K+1}$. This is defined in our case as follows, with $a$ equal to *pickup*$(\text{A}, \text{B})$ and $s'$ ranging over possible next states:

$$\sum_{s'} \left| p(s'|s, a, R_{K+1}) - p(s'|s, a, \widehat{R}_{K+1}) \right|$$

   Finally, compute the average variational distance over the test set.

Variational distance is a measure of error, but we would like the y-axis in our graphs to be a measure of accuracy, so we use $1 - (\text{variational distance})$.

The free parameters in our hierarchical Bayesian model (and hence in our scoring function) are set to the same values in all experiments. While we found that the scoring function in Eq. 11 leads to good results on large training sets, we also saw that with small training sets, the very small probabilities of formulas (in contexts and outcomes) tend to dominate the score. For the experiments reported here, we use a modified scoring function in which each occurrence of the formula distribution $P_{\text{for}}$ is raised to the power 0.5. The fact that this *ad hoc* modification yields better results suggests that our distribution over formulas is overly flat, and it would be worthwhile to develop a formula distribution that gives common literals or subformulas higher probability.

[5] The distribution used here is biased so that A is always a block and the robot's gripper is usually empty; this focuses our evaluation on cases where *pickup*$(\text{A}, \text{B})$ has a chance of success.

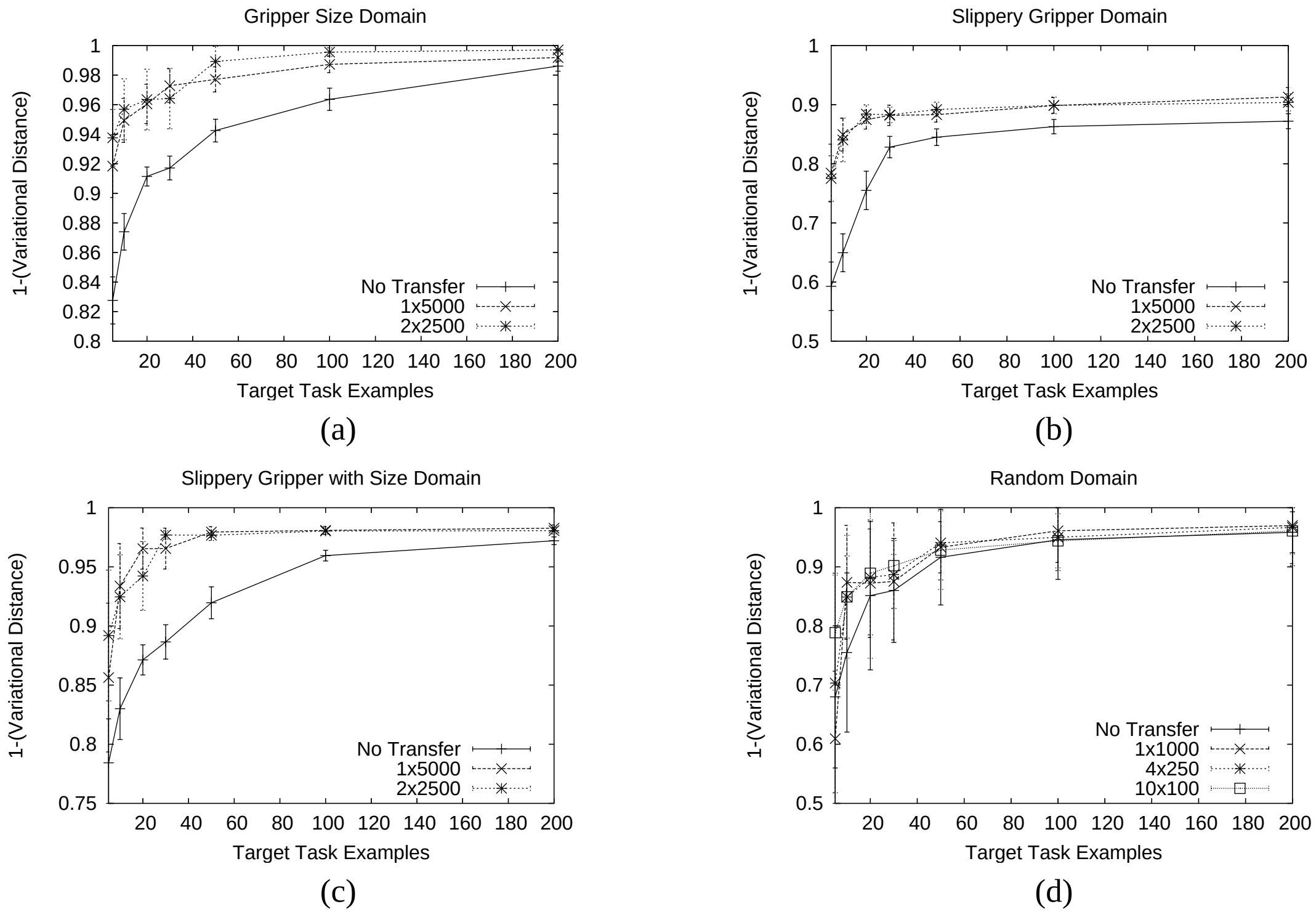

Figure 3: Accuracy using an empty rule set prototype (labeled "No Transfer") and transfer learning, labeled $K$x$N$ where $K$ represents the number of source tasks and $N$ represents the number of examples per source task.

## 5.2 Results

In this section, we present results in the four blocks world domains. For each domain, we briefly describe the task generation distribution and then present results.[6] For each experiment, we graph variational distance as a function of the number of training examples in the target task. Each experiment was repeated 20 times; our graphs show the average results with 95% confidence bars. The time required for each run varied from 30 seconds to 10 minutes depending on the complexity of the domain.

Our first experiment investigates transfer learning in a domain where the rule sets are very simple — just single rules — but the rule contexts vary across tasks. We use a family of tasks where the robot is equipped with grippers of varying sizes. There are seven different sizes of blocks on the table; the robot can only pick up blocks that are the same size as its gripper. Thus, each task can be described by a single rule saying that if block $X$ has the proper size, then *pickup*$(X, Y)$ succeeds with some significant probability (this probability also varies across tasks). If $X$ has the wrong size, then no rule applies and there is no change. Since the "proper size" varies from task to task, the rules for different tasks have different contexts. To increase the learning difficulty, two extra distracter predicates (*color* and *texture*) are randomly set to different values in each example state.

Fig. 3(a) shows the transfer learning curves for this domain. The transfer learners are consistently able to learn the dynamics of the domain with fewer examples than the non-transfer learner. In practice, in each source task, the algorithm learns the specific *pickup* rule with the appropriate *size* literal in the context. The algorithm learns a single rule prototype whose context also contains some *size* literal. This rule prototype provides a strong bias for learning the correct target-task rule set: the learner only has to replace the *size* literal in the prototype with the correct *size* literal for the given task.

To see how transfer learning works for more complex rule sets, our next experiment uses a "slippery gripper" domain adapted from [Kushmerick *et al.*, 1995]. The correct model for this domain has four fairly complex rules, describing cases where the gripper is wet or not wet (which influences the success probability for *pickup*) and the block is being picked up from the table or from another block (in the latter case, the rule must include an additional outcome for the block falling on the table). The various tasks are all modeled by rules with the same structure, but include relatively large variation in outcome probabilities.

[6]Deshpande [2007] presents a more detailed description of these domains.

Fig. 3(b) shows the transfer learning curves for the slippery gripper domain. Again, transfer significantly reduces the number of examples required to achieve high accuracy. We found that the transfer learners create prototype rule sets that effectively represent the dynamics of the domain. However, the structure of the prototype rules do not exactly match the structure of the four specific rules that are present in each source task. Despite this fact, these prototypes still capture common structure that can be specialized to quickly learn the correct rules in the target task.

Our third domain, the slippery gripper domain with size, is a cross between the slippery gripper domain and the gripper size domain. In this domain, all four rules of the slippery gripper domain apply with the addition that each rule can only succeed if the targeted block is of a certain task-specific size. Thus, the domain exhibits both structural and parametric variation between tasks.

As can be seen in Fig. 3(c), the transfer learners perform significantly better than the non-transfer learner. In this case, the rule set prototype provides both a parametric and structural bias to better learn the domain.

Our final experiment investigates whether our algorithm can avoid erroneous transfer when the tasks are actually unrelated. For this experiment, we generate random source and target rule sets with 1 to 4 rules. Rule contexts and outcomes are of random length and contain random sets of literals. Since rule sets sampled this way may contain overlapping rules or outcomes, we use rejection sampling to ensure that a valid rule set is generated for each task.

As can be seen in Fig. 3(d), the transfer and non-transfer learners' performances are statistically indistinguishable. The learning algorithm often builds a rule set prototype containing a few rules with random structure and high variance outcome distribution priors. These prototype rules do not provide any specific guidance about the structure or parameters of the specific rules to be learned in the target task. However, their presence does not lower performance in the target task.

## 6 Conclusion

In this paper, we developed a transfer learning approach for relational probabilistic world dynamics. We presented a hierarchical Bayesian model and an algorithm for learning a generic rule set prior which, at least in our initial experiments, holds significant promise for generalizing across different tasks. This learning problem is particularly difficult due to the need to learn relational structure along with probabilities simultaneously for a large number of tasks. The current approach addresses many of the fundamental challenges for this task and provides a strong example that can be extended to work in more complex domains and with a wide range of representation languages.